\documentclass[fleqn,10pt,twocolumn]{AROB_TexTemplate/AROB}
\usepackage[square,numbers]{natbib}

\usepackage{amsfonts} \usepackage{amssymb} \usepackage{amsmath}
\usepackage{xcolor}
\usepackage{booktabs}

\title{Real-Time Motion Planning with Dynamic Hazards: \\
        Classical vs. Learning-Based Methods
}

\author{Eran Iceland${}^{1,*}$, Alexander Tuisov${}^{2,*}$, Oren Gal${}^{3}$, Ariel Barel${}^{4\dagger}$, and Alfred M. Bruckstein${}^{4}$}

\speaker{Ariel Barel}

\affils{${}^{1}$School of Engineering and Computer Science, The Hebrew University of Jerusalem, Jerusalem, Israel\\
(E-mail: eran.iceland@gmail.com)\\
${}^{2}$Faculty of Data and Decision Science, Technion Israeli Institute of Technology, Haifa, Israel\\
(E-mail: queldelan@gmail.com)\\
${}^{3}$Hatter Department of Marine Technologies, University of Haifa, Haifa, Israel\\
(E-mail: orengal@univ.haifa.ac.il)\\
${}^{4}$Faculty of Computer Science, Technion Israeli Institute of Technology, Haifa, Israel\\
(E-mail: arielba@technion.ac.il; freddy@cs.technion.ac.il)\\
${}^{*}$These authors contributed equally to this work.
}
\abstract{%
We study real-time motion planning in dynamic hazard fields through a controlled comparison between classical planning and learning-based methods. Rather than introducing a new planner, we construct a unified benchmark in which representative classical and learning-based methods face the same environments, motion constraints, information assumptions, and evaluation metrics. The test environment consists of planar domains populated with rotating sprinkler-like hazards that generate time-varying forbidden regions via sweeping angular sectors.
Our results show a clear regime shift. In deterministic environments, classical planners achieve near-perfect success and higher-quality paths, though sometimes at the cost of substantial planning or replanning time. Under stochastic obstacle dynamics, however, online search becomes strongly budget-sensitive: low budgets lead to frequent failure, while high budgets improve success at the cost of latency and longer trajectories. PPO-based policies, trained under the same scenario distribution, consistently outperform in latency, success rate, and path quality in these stochastic regimes. Overall, the results indicate that uncertainty in obstacle evolution, more than partial observability, is the dominant factor determining which planning paradigm is practically effective for the problem at hand.}

\begin{document}

\maketitle


\begin{center} \small This paper is scheduled to be presented at the 2027 International Symposium on Artificial Life and Robotics (AROB 2027). \end{center}


\section{Introduction}

We consider motion planning in planar environments with dynamic hazards, a setting that has been extensively studied through both classical and learning-based approaches. Classical methods provide precise geometric or safety-driven guarantees under explicit modeling assumptions, while recent Reinforcement Learning (RL)-based planners offer adaptability and scalability in partially known or stochastic environments. Despite substantial progress on each side, relatively few works provide a controlled, side-by-side comparison between classical and RL-based methods under consistent modeling assumptions and evaluation criteria. In contrast to prior work that introduces new algorithms, our goal is to enable a controlled comparison under a shared problem formulation, isolating regime-dependent performance differences rather than architectural novelty, while providing practical guidance on when each paradigm should be preferred.\\
To achieve that, we construct a unified testbed that is simple enough to admit principled classical treatment, yet rich enough to expose the limitations of each paradigm. We begin with the simplest regime, in which the full evolution of the environment is known in advance, and progressively introduce complexity along two axes: visibility (full vs. limited) and obstacle dynamics (deterministic vs. stochastic). This yields four regimes, ranging from globally known, deterministic settings to partially observable and randomly evolving environments, enabling a systematic comparison under controlled conditions. This design choice intentionally prioritizes comparability over novelty in algorithm design, allowing us to attribute observed performance differences to problem structure rather than to formulation mismatches.\\
We formalize the problem setting as follows. Under full visibility, we consider a deterministic formulation via an $XYT$ lifting \citep{AdinBruckstein1991, KimmelKiryatiBruckstein1998, fujimura2002time}.
The robot is modeled as a holonomic point moving at fixed speed and bounded turn rate in the plane while time advances monotonically, i.e. the agent must advance monotonically along \(T\) at unit rate. Dynamic obstacles make feasibility time-dependent, creating a tight coupling between spatial motion and temporal progression.


\subsection{2D/3D Static and Dynamic Planning}
\label{sec:analytic-2d3d}
In 2D with static obstacles, classical path planning commonly relies on geometric constructs such as visibility graphs, Voronoi-based roadmaps, or search-based methods such as A* over discretized grids \citep{LozanoPerezWesley1979,Latombe1991,Mitchell2000ShortestPaths,Choset2005}. The setting is simpler because the environment is fixed; the main difficulty is a continuous-space combinatorial search. When obstacles are dynamic in 2D, the search naturally lifts to the spacetime domain \(XYT\): feasibility must hold simultaneously in space and time, increasing dimensionality and introducing temporal constraints. Early work on real-time navigation in dynamic environments addressed these temporal constraints through analytic, safety-driven online planners.
In particular, velocity-obstacle-based approaches explicitly avoid states of inevitable collision by selecting admissible motions over a carefully chosen time horizon \cite{GalShillerRimonICRA2009}.
Subsequent work introduced adaptive time-horizon selection, reducing conservatism while preserving online safety guarantees in crowded and time-varying scenes \cite{ShillerGalRazIROS2011}. Classical tools in such settings include time-expanded graphs originating from dynamic-flow formulations \citep{FordFulkerson1958}, reachability-based methods grounded in Hamilton-Jacobi optimal control analysis \citep{MitchellBayenTomlin2005}, and reactive models such as velocity obstacles or the dynamic window approach \citep{FioriniShiller1998,Fox1997}.

In 3D with static obstacles, geometric complexity rises: obstacle representations, distance queries, and feasibility checks become more expensive. Sampling-based planners such as PRM and RRT/RRT* are widely used for high-dimensional, non-convex environments \citep{Kavraki1996,LaValleKuffner2001,KaramanFrazzoli2011,LaValle2006}. With dynamic obstacles in 3D, the problem extends to \(XYZT\), introducing non-convex and often non-linear planning challenges that require predictive obstacle-motion models and uncertainty-aware reasoning \citep{Thrun2005}.

\subsection{Reinforcement Learning for Obstacle Avoidance}
Recent surveys of deep RL for navigation in dynamic and crowded environments highlight strong empirical promise, while emphasizing persistent challenges in generalization and real-world deployment \citep{Zhu2025DRLDynamicReview,Le2024CrowdedDRLReview}.
They also note that results are often difficult to compare across papers due to heterogeneous assumptions about sensing, obstacle dynamics, and safety constraints \citep{AbuJabal2024DynamicPathPlanningSurvey}.

These observations motivate our controlled evaluation: we compare classical and RL-based planners under consistent assumptions and metrics to isolate regime-dependent performance differences.


We focus on the 2D dynamic-obstacle setting because it allows a consistent treatment across all regimes of interest while remaining computationally tractable. Under full visibility, the problem admits principled solutions that leverage the known evolution of the environment, including spacetime formulations and asymptotically optimal planning methods \citep{KaramanFrazzoli2011}. Under limited visibility, the same formulation naturally leads to incremental and online replanning strategies as information is revealed during execution \citep{Stentz1995,KoenigLikhachev2002}. This duality makes the setting particularly well-suited for systematic evaluation of both classical and RL-based planners within a unified problem formulation.

\subsection{Contribution Statement}

 Our contributions are primarily empirical and methodological. We introduce a unified testbed for motion planning in dynamic environments, enabling a controlled comparison between classical and RL-based methods under consistent modeling assumptions. The testbed spans four world models defined by two axes: visibility, either full or limited, and obstacle dynamics, either deterministic or stochastic. For these regimes, we implement classical baselines tailored to the available information and dynamics assumptions, together with a consistent RL formulation using a geometrically interpretable action space and compatible observation models. We then provide an empirical evaluation comparing the resulting planners in terms of success rate, path quality, computational cost, and robustness to uncertainty.

 Our objective is not to model a specific robotic platform or deployment scenario, but rather to isolate the effect of information availability and obstacle uncertainty on planning performance within a controlled experimental framework.

\section{Preliminaries and Notation}
\label{sec:notation}

\begin{itemize}
    \item Workspace: $W \subset R^2$ bounded.

    \item Dynamic obstacles: $O(t)=\{O_i(t)\}_{i=1}^m$, with each $O_i(t) \subset W$ closed. Free space is denoted by $F(t)$. In what follows, the dynamic obstacles are instantiated as rotating circular sectors, which we refer to as sprinklers.

    \item State kinematics: position $(x_t,y_t)$, heading $\theta_t \in [-\pi,\pi)$, time $T_t \ge 0$.

    \item Start and goal: $s=(x_s,y_s)$, $g=(x_g,y_g)$. The agent is considered to have reached the goal if $\|(x_t,y_t)-g\|<1$.

    \item Constant-speed constraint: $\|(x_{t+1},y_{t+1})-(x_t,y_t)\|=1$ (discrete time). Time advances monotonically at unit rate.

    \item A discrete action space of $7$ steering actions:
    \begin{equation}
    \label{eq:action_space}
    a_t = \Delta \theta_t \in
    \{-45^\circ,-30^\circ,-15^\circ,0^\circ,15^\circ,30^\circ,45^\circ\}.
    \end{equation}

\end{itemize}
\subsection{Problem Regimes} \label{sec:info-models} 
\begin{figure}[t]
  \centering
\includegraphics[width=\linewidth]{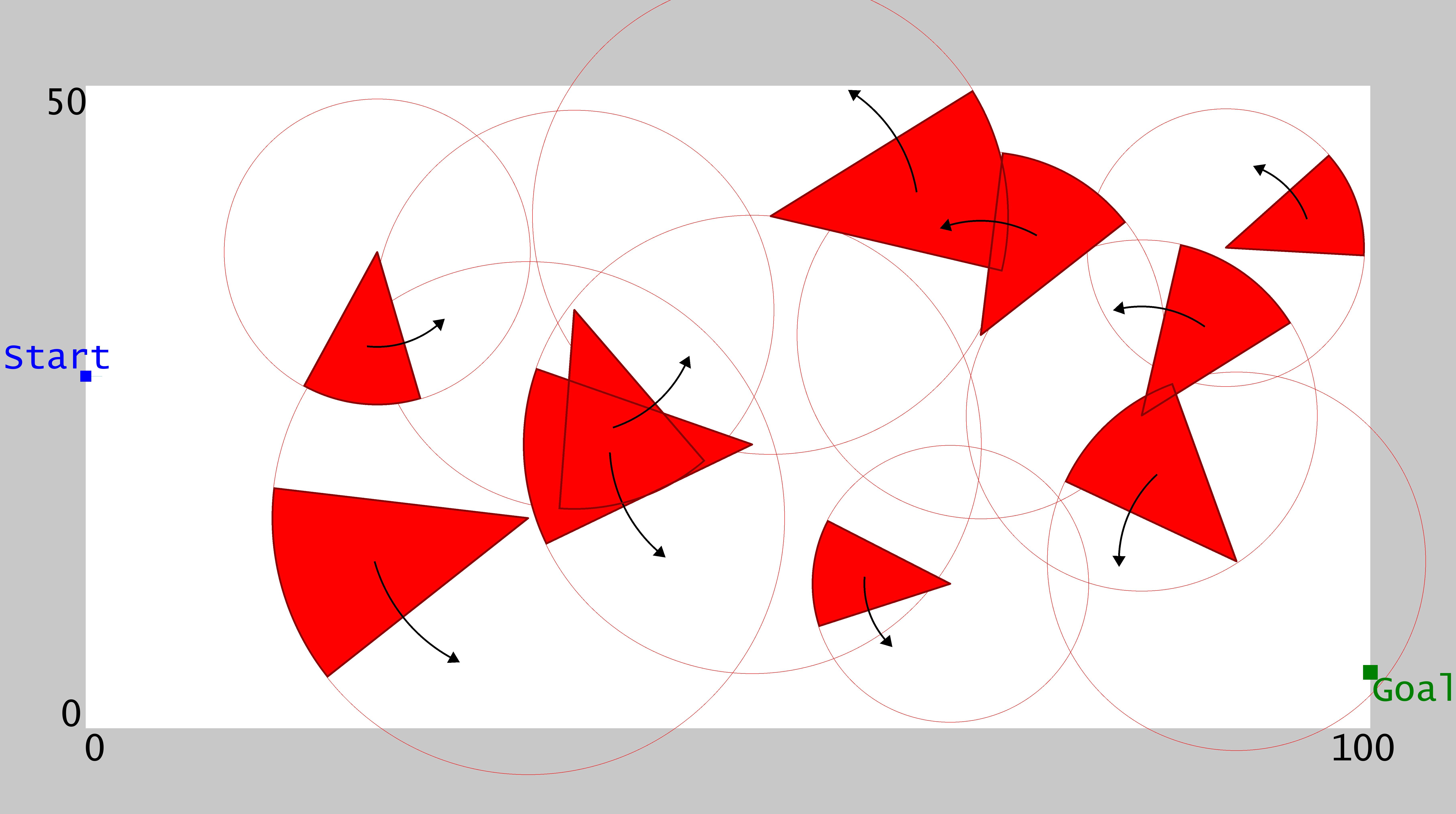}
\caption{Illustration of the environment used throughout the experiments: a rectangular playground with rotating ``sprinklers'' acting as dynamic hazards. Each sprinkler generates a circular sector that rotates, inducing time-varying forbidden regions. The agent starts on one side of the domain and seeks a shortest-time trajectory to the opposite side while remaining collision-free (dry), subject to constant-speed and bounded-turn-rate motion constraints.}
  \label{fig:sprinklers}
\end{figure}

\begin{figure*}[t] \centering \includegraphics[width=0.48\textwidth]{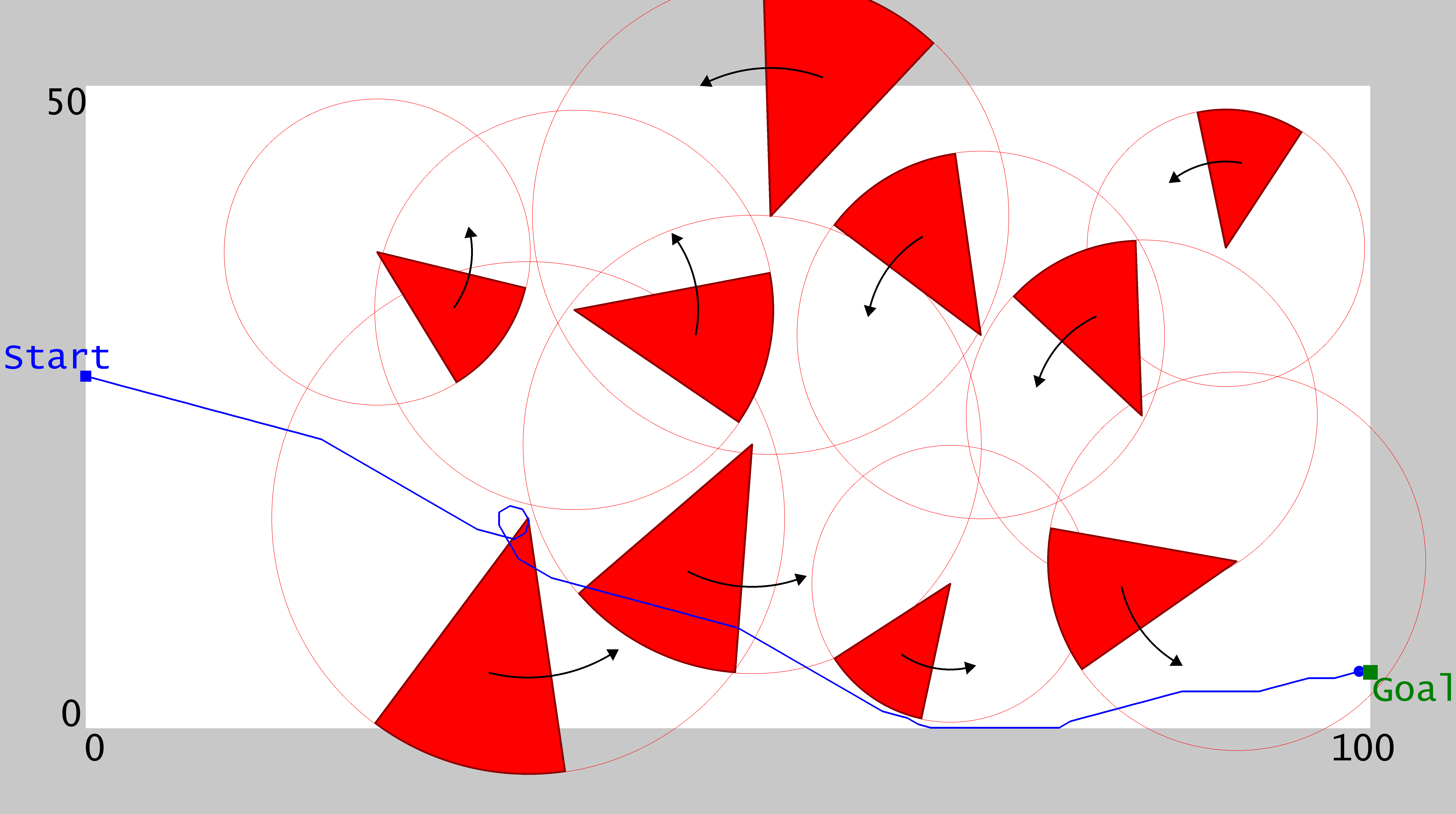} \hfill \includegraphics[width=0.48\textwidth]{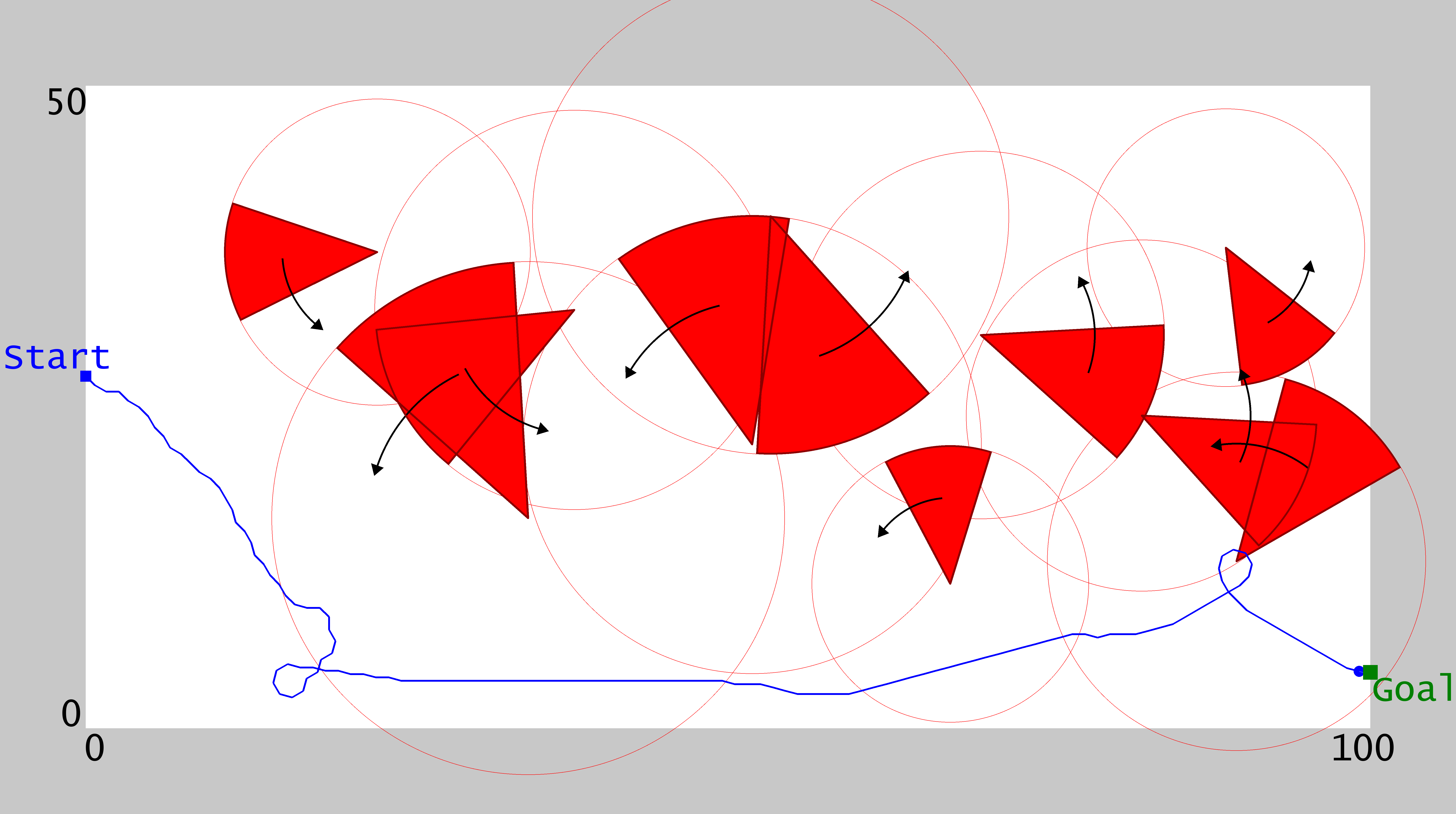} \caption{ Left: A* (precomputed). Right: RL (real-time). Valid trajectories under full visibility with deterministic dynamics.} 
\vspace{4pt}
\label{fig:sprinklers-trajectories} \end{figure*}

We consider four regimes obtained by combining two axes within the above formalization. Along the visibility axis, \emph{full visibility} assumes that the obstacle set ${\mathcal O}(t)$, and hence the free space ${\mathcal F}(t)$, is known for all $t$, whereas \emph{limited visibility} restricts the agent to partial, local observations of ${\mathcal F}(t)$ at each time $t$ around agent's location at $t$. Along the dynamics axis, \emph{constant angular velocity} obstacles correspond to deterministic trajectories $O_i(t)$ with fixed motion parameters, while \emph{stochastically varying angular velocity} allows these trajectories to evolve randomly over time. This yields four planning regimes: (i) full visibility with constant angular velocity obstacles, where ${\mathcal O}(t)$ is fully known and deterministic; (ii) limited visibility with constant angular velocity obstacles, where ${\mathcal O}(t)$ is deterministic but only partially observed; (iii) full visibility with stochastically changing angular velocities, where ${\mathcal O}(t)$ evolves randomly but is globally observable; and (iv) limited visibility with stochastic dynamics, where ${\mathcal O}(t)$ is both partially observed and randomly evolving. In all cases, feasibility is defined with respect to ${\mathcal F}(t)$ satisfying \( \|(x_{t+1}, y_{t+1}) - (x_t, y_t)\| = 1 \). The visualization of the initial state can be seen in Figure~\ref{fig:sprinklers}, while the visualization of the corresponding trajectories can be seen in Figure~\ref{fig:sprinklers-trajectories}.

\section{Classical Search and Planning Baselines}
\label{sec:analytic}

For each regime in Section~\ref{sec:info-models}, we implement a classical baseline matched to the regime's information and dynamics assumptions. These baselines are representative methods for controlled comparison with the RL planner.

\subsection{State-Space Search Formulation}

A state-space search problem is defined by an initial state, a successor function, a goal test, and a path-cost function \citep{russell2016artificial}. We use states $(x,y,\theta,t)$, where $(x,y)$ is the agent position, $\theta$ its heading, and $t$ the timestep. Actions follow the motion constraints in Section~\ref{sec:notation}; successors are generated by applying feasible actions, and the goal test checks whether the agent is within one step of the goal.

\subsection{Deterministic Obstacle Angular Velocity}

With constant obstacle angular velocity, the future hazard field is deterministic once the scenario is specified, so planning reduces to graph search over the discretized spacetime state space $(x,y,\theta,t)$. Exact continuous planning is difficult in dynamic domains: while static 2D shortest paths can be handled by visibility graphs or exact cell decompositions \citep{Latombe1991,Mitchell2000ShortestPaths}, lifting to \(XYT\) breaks the usual vertex-only structure, and dynamic alternatives such as velocity-obstacle or dynamic-window methods still require online replanning \citep{Mitchell2000ShortestPaths,FioriniShiller1998}.

We use A* \citep{hart1968formal} as the high-quality deterministic baseline: with an admissible heuristic, it is complete and optimal with respect to the induced discretized graph, though not necessarily the original continuous problem. We also evaluate Greedy Best-First Search (GBFS) with the same successor model and heuristic.  GBFS is not optimal, but provides a faster satisficing baseline. For both A* and GBFS, we use Euclidean distance to the goal as the heuristic. Together, A* and GBFS expose the standard speed-optimality tradeoff under deterministic obstacle evolution.

\subsubsection{Limited Visibility} Under limited visibility, the planner observes only part of the spacetime graph, which can be discovered during execution. We therefore use a D*-like replanning framework \citep{Stentz1995}: plan on the currently observed map, execute forward, and repair the plan when newly observed sprinklers invalidate it. In this setting, A* and GBFS are online replanning components rather than globally optimal planners. A* remains optimal only for the currently known planning instance, and neither method is complete with respect to the full hidden environment: a feasible full-information trajectory may exist even if the online agent later reaches a deadlock.

\subsection{Stochastic Obstacle Angular Velocity}

With stochastic obstacle angular velocities, the future hazard field is no longer a single known trajectory. Deterministic shortest-path methods such as A* are therefore not directly applicable without replacing the stochastic process by a fixed prediction, which fundamentally undermines robustness to uncertainty.

We use Upper Confidence bounds applied to Trees (UCT) \citep{kocsis2006bandit} as a stochastic planning baseline. UCT performs sampled lookahead over possible future obstacle evolutions, balancing exploration and exploitation in the search tree. At each decision step, branches correspond to action sequences and sampled obstacle trajectories; leaf nodes are evaluated using Euclidean distance from the goal rather than full Monte Carlo rollouts. We use frame stacking \citep{mnih2015human}, selecting actions every \(k=5\) timesteps to increase the effective lookahead horizon.

UCT is run as an anytime planner with a fixed computation budget per decision step. This budget controls the amount of stochastic lookahead, making UCT suitable for measuring the runtime-quality tradeoff of planning under uncertainty. While not exhaustive of all stochastic planners, UCT is a principled and widely used baseline for online decision-making in stochastic environments.

\section{RL-Based Planner}
\label{sec:rl}

We model the planner as an episodic RL agent executed in closed loop. At each timestep, the agent observes the environment, selects an action, advances one step, and receives a reward. Policies are trained using PPO \citep{schulman2017proximal}, chosen for its stability under sparse and delayed feedback in dynamic environments and its common usage in image-based control with discrete action spaces, using the action space defined in Equation~(\ref{eq:action_space}). This enables direct comparison with classical planners. The dynamics follow unit-speed kinematics, i.e., the agent moves one unit per timestep in the direction of its heading.

Evaluation is performed on scenarios sampled from the same distribution as training, and policies are executed without further learning or adaptation. Full implementation details of the RL agent, including observations, reward design, network architecture and training procedure, are deferred to Appendix~\ref{app:rl_details}.

\section{Experiments and Results}
\label{sec:experiments}
We evaluate classical and RL-based methods under all four world models described in Section \ref{sec:info-models}. As a classical method we evaluate appropriate methods described in Section \ref{sec:analytic}. For RL methods we trained policies described in Section \ref{sec:rl}.


\subsection{Experimental Setup}\label{Experimental Setup}
We ran 1,000 randomly generated inputs for each world model, and ran an RL-based and an appropriate classical algorithm on it. Scenarios were generated using the following protocol: 

\begin{itemize}
    \item The workspace is a \(100 \times 50\) rectangle; agents start at \(x=0\) with random \(y\) and fixed eastward heading, and goals are placed at \(x=100\) with random \(y\).
    \item Sprinkler positions were sampled using a Greedy Max-Min rule \citep{gonzalez1985clustering}, each time selecting the location farthest from the current set to improve worst-case coverage, i.e., $x=\arg\max_{p}\min_{s\in S} d(p,s)$, ensuring non-degenerate and well-spread hazard configurations, with centers constrained to satisfy \(22 < x < 90\) to avoid trivially unsolvable instances. 
    \item  Radii were sampled uniformly from a prescribed interval \(R \in [10,20]\).
    \item Angular velocities: in the constant angular velocity regime, fixed at \(10^\circ/\text{s}\) (\(\pi/18\) rad/s), counterclockwise; in the stochastic regime, updated via a clipped and discretized random walk \( \omega_{t+1} = \omega_t + \epsilon_t \), where \(\epsilon_t \sim \mathrm{Normal}\!\left(0,\left(0.5^\circ\right)^2\right)\), values are quantized to increments of \(0.25^\circ\), and constrained to the interval \(5^\circ/\text{s} \leq \omega_t \leq 15^\circ/\text{s}\).
    \item For the limited visibility setting, we used vision radius $V=20$ 
    \item To ensure a fair comparison, all algorithms were evaluated on identical scenarios generated using a fixed random seed, with a common cutoff horizon of 400 steps.
    \item For hardware details, including our controlled setup ensuring identical execution conditions across all experiments, see Appendix~\ref{app:hardware}.

\end{itemize}


\subsection{Quantitative Results}
\label{sec:results}

\begin{table*}[tbp] \centering \caption{Performance under constant obstacle angular velocity based on 1,000 runs.} \label{tab:constant_speed} \normalsize \setlength{\tabcolsep}{8pt} \renewcommand{\arraystretch}{1.25} \begin{tabular}{llcccc} \hline \textbf{Vis.} & \textbf{Alg.} & \textbf{Succ. (\%)} & \textbf{Path Length} & \textbf{Offline Time (s)} & \textbf{Max Replan Time (s)} \\ \hline Full & A* & \textbf{100.0} & \textbf{114.38 (8.06)} & 38.56 (40.29) & NA \\ Full & GBFS & 100.0 & 137.41 (21.59) & 11.27 (3.43) & NA \\ Full & RL GPU & 85.3 & 135.06 (21.69) & NA & \textbf{0.0017 (0.0003)} \\ Full & RL CPU & 85.3 & 135.06 (21.69) & NA & 0.0027 (0.0004) \\ \hline Limited & A* & \textbf{96.4} & \textbf{117.12 (10.50)} & NA & 5.7458 (7.8757) \\ Limited & GBFS & 92.2 & 137.44 (22.50) & NA & 2.2289 (1.3642) \\ Limited & RL GPU & 81.3 & 136.15 (22.21) & NA & \textbf{0.0016 (0.0002)} \\ Limited & RL CPU & 81.3 & 136.15 (22.21) & NA & 0.0027 (0.0004) \\ \hline 
\hline \end{tabular} \vspace{8mm} \end{table*}
\begin{table*}[tbp] 
\centering \caption{Performance under stochastic obstacle angular velocity based on 1,000 runs.} \label{tab:variable_speed} \normalsize \setlength{\tabcolsep}{8pt} \renewcommand{\arraystretch}{1.25} \begin{tabular}{llccc} \hline\hline \textbf{Vis.} & \textbf{Alg.} & \textbf{Succ. (\%)} & \textbf{Path Length} & \textbf{Max Replan Time (s)} \\ \hline Full & UCT (0.002s) & 3.9 & 136.92 (27.07) & 0.0020 (0) \\ Full & UCT (0.01s) & 14.0 & 121.26 (11.91) & 0.0100 (0) \\ Full & UCT (0.1s) & 52.1 & 182.57 (57.89) & 0.1000 (0) \\ Full & UCT (1s) & 75.6 & 198.75 (62.80) & 1.0000 (0) \\ Full & RL GPU & \textbf{76.3} & \textbf{152.26 (31.78)} & \textbf{0.0016 (0.0002)} \\ Full & RL CPU & 76.3 & 152.26 (31.78) & 0.0027 (0.0002) \\ \hline Limited & UCT (0.002s) & 4.1 & 134.63 (19.27) & 0.0020 (0) \\ Limited & UCT (0.01s) & 13.8 & 121.02 (11.97) & 0.0100 (0) \\ Limited & UCT (0.1s) & 53.4 & 179.42 (53.79) & 0.1000 (0) \\ Limited & UCT (1s) & 74.8 & 199.57 (62.64) & 1.0000 (0) \\ Limited & RL GPU & \textbf{75.9} & \textbf{150.48 (32.76)} & \textbf{0.0017 (0.0003)} \\ Limited & RL CPU & 75.9 & 150.48 (32.76) & 0.0027 (0.0003) \\ \hline\hline \end{tabular} \end{table*}
In the deterministic (constant angular velocity) regime presented in Table \ref{tab:constant_speed}, classical planning methods clearly outperform RL-based approaches, particularly under full visibility. Both A* and GBFS achieve perfect success rates, with A* producing near-optimal trajectories and GBFS trading optimality for speed. Even under limited visibility, classical methods remain robust, exhibiting only moderate degradation (from 100\% to approximately 96\% success for A*), which reflects the effectiveness of incremental replanning strategies. In contrast, RL methods achieve consistently lower success rates (around 80-85\%) and do not provide a compensating advantage in path quality. These results indicate that when the environment is deterministic, even if only partially observed, classical planners can effectively exploit structural information and maintain high performance. When per-decision latency is the primary constraint, the RL policy provides a much faster online alternative, at the cost of lower success and path quality. A* offline calculation time is heavy-tailed; see Appendix~\ref{app:astar_offline_hist}.

\begin{figure*}[t]
\centering

\begin{minipage}[t]{0.44\textwidth}
\centering
\includegraphics[width=\linewidth]{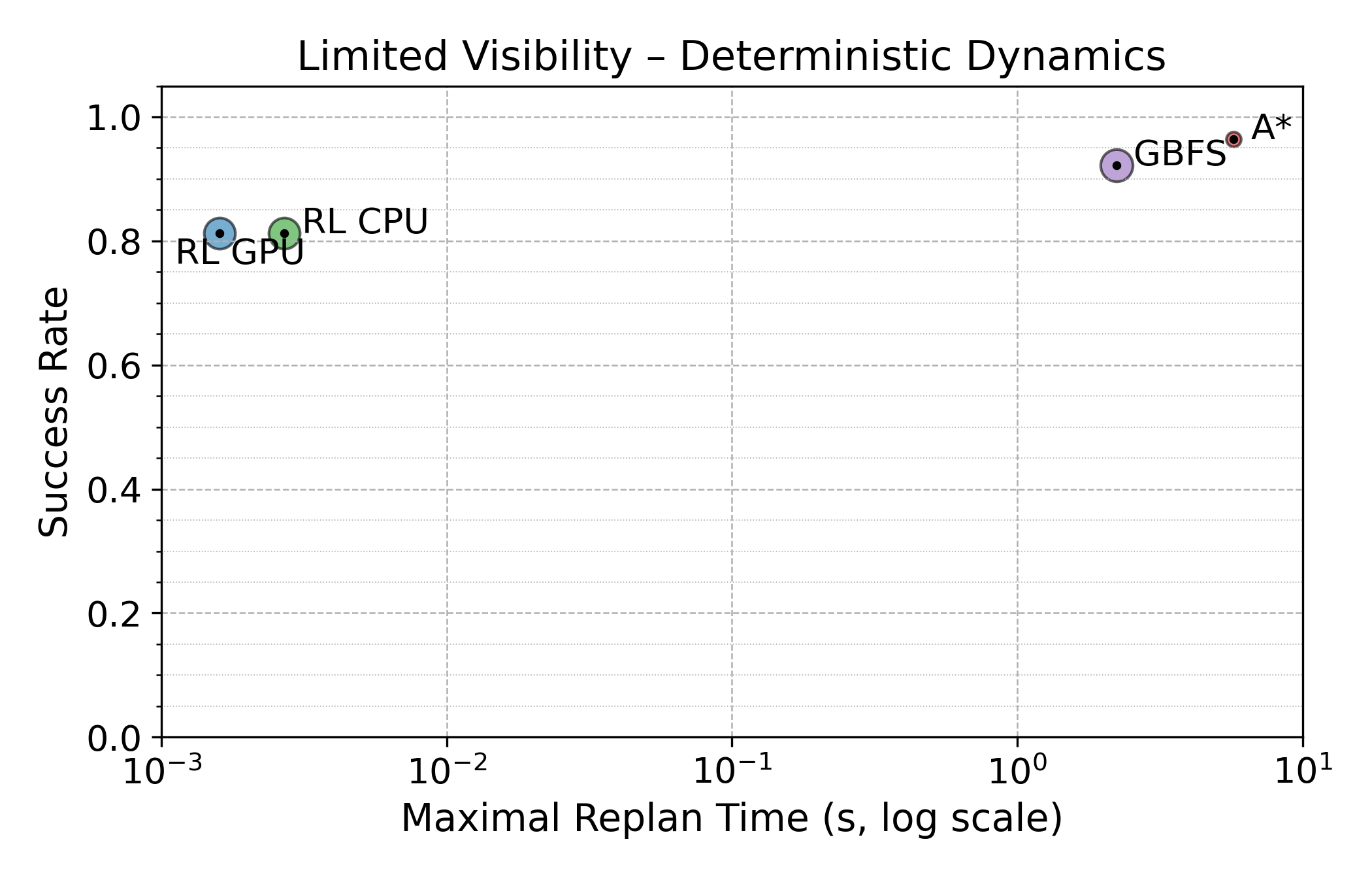}

(a) Deterministic
\end{minipage}
\hfill
\begin{minipage}[t]{0.44\textwidth}
\centering
\includegraphics[width=\linewidth]{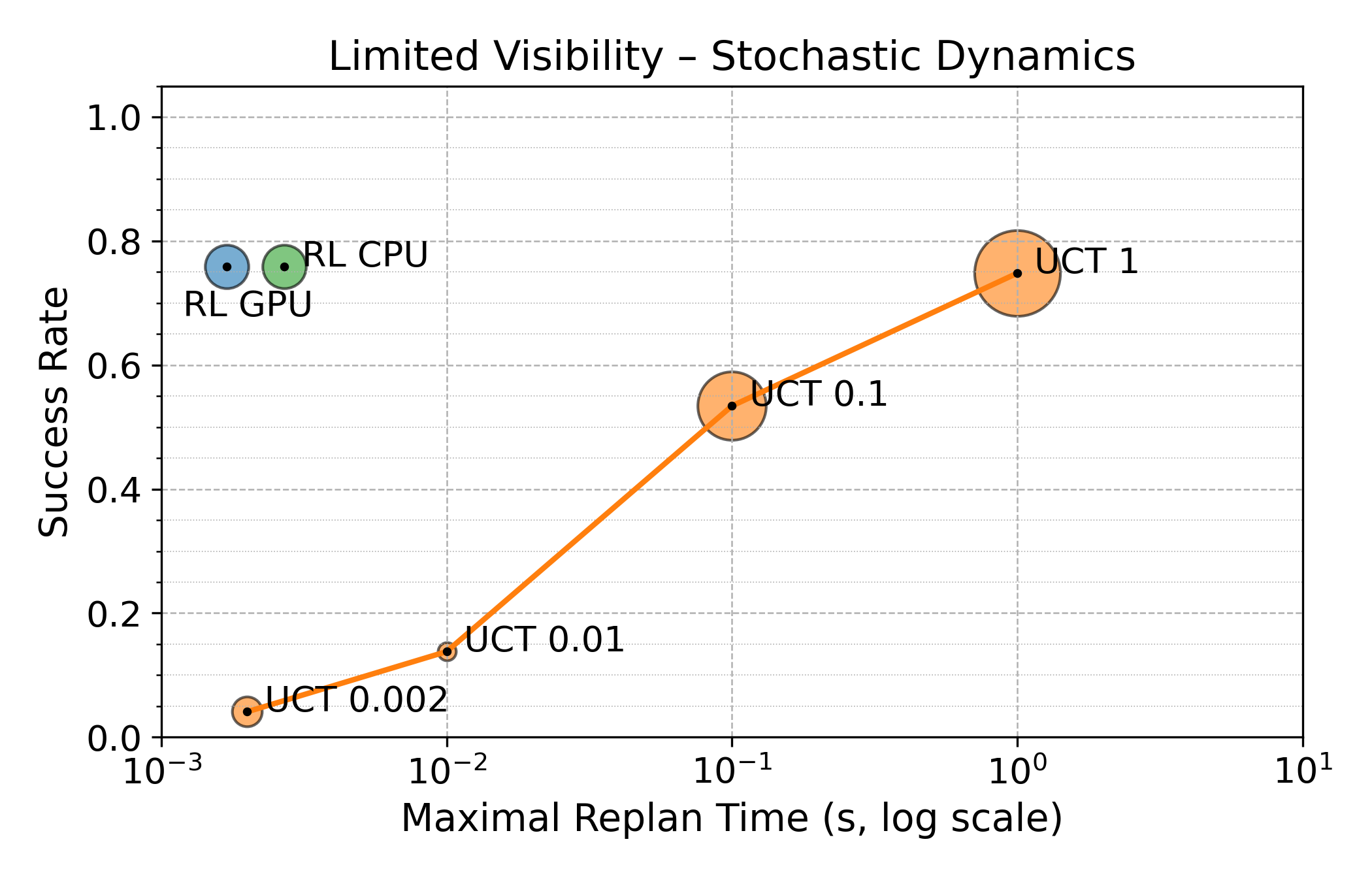}

(b) Stochastic (budgeted UCT)
\end{minipage}

\caption{Success rate versus maximal replan time under limited visibility for (a) deterministic and (b) stochastic dynamics. Marker size represents path length; the UCT curve connects different per-step planning budgets.}
\label{fig:sprinklers_conclusion_diagram}
\end{figure*}

The introduction of stochastic obstacle dynamics fundamentally alters this picture as can be seen in Table \ref{tab:variable_speed}. Classical methods, represented by UCT with varying computational budgets, exhibit strong sensitivity to the available planning time: small timeouts lead to near-complete failure, while larger budgets improve success rates but result in significantly longer and more conservative trajectories. In contrast, RL methods achieve stable performance across regimes, matching or slightly exceeding the best UCT configurations in success rate while maintaining substantially shorter paths and low computational cost. Importantly, this degradation of classical methods is driven by uncertainty rather than limited visibility, as performance drops sharply even under full visibility once stochastic dynamics are introduced. This demonstrates that uncertainty in obstacle evolution is the dominant factor affecting planning difficulty, and that the trained RL policy consistently outperforms our online UCT baseline under stochastic obstacle evolution in latency, success rate, and path quality. This comparison is visualized in Fig.~\ref{fig:sprinklers_conclusion_diagram}.

\subsection{Ablation Study}
\label{sec:ablations}

\subsubsection{Classical Planners.}
For the classical stochastic planner, the main sensitivity was the UCT decision interval.
Under full visibility with a 1s budget, frame stacking with \(k=5\) achieved \(75.6\%\)
success, compared with \(15.6\%\) for \(k=1\), indicating that effective lookahead depth is
critical under stochastic dynamics. We also tested a risk-aware heuristic and alternative
discretizations, but preliminary results were not competitive, so these variants were not run
across the full suite.

\subsubsection{RL Methods.}

We evaluated the sensitivity of the learned planner to key training and reward parameters. Varying the discount factor within \([0.95, 0.999]\) had no noticeable effect, suggesting limited sensitivity to long-horizon credit assignment in this setting. Similarly, adding simple shaping terms, including a per-step penalty of -\(0.01\) or a rotation penalty of -\(0.01|\Delta\theta|\), did not yield measurable improvements. In contrast, changing the balance between sparse and dense rewards had a clear effect: reducing the completion reward from \(25\) to \(5\) while increasing the progress reward from \(0.005\) to \(0.1\) shortened paths under full visibility with fixed sprinkler angular velocity from \(135\) to \(127\) steps, but reduced success under full visibility with deterministic angular velocity from \(85\%\) to \(75\%\). This suggests that stronger dense rewards encourage locally efficient motion, whereas a larger terminal reward is important for reliable task completion.

\section{Limitations}
\label{sec:limitations}

The comparison should be interpreted within the scope of our benchmark. The environment
is a structured 2D domain with simplified sprinkler dynamics, discrete steering actions,
constant-speed motion, and perfect execution. The hazards are non-adversarial and follow
simple rotational dynamics; richer robotic settings may involve sensing noise, actuation
error, dynamic agents with strategic behavior, non-holonomic dynamics, or more complex
geometry. Moreover, the RL policies are evaluated in-distribution: test scenarios are sampled
from the same scenario family used during training, so the results do not establish broad
out-of-distribution generalization.

Our evaluation also focuses on online performance. We report planning and inference
latency during execution, but do not include the offline cost of PPO training. Conversely,
the stochastic classical baseline is represented by UCT with fixed per-step budgets; other
model-based approaches, such as sampling-based MPC, belief-space planning, or
chance-constrained planning, may yield different tradeoffs. Finally, our metrics emphasize
success, path length, and latency. Future work should include richer safety criteria, such as
near-miss rates, accumulated risk exposure, and robustness under shifted obstacle
distributions.

\section{Conclusion}
\label{sec:conclusion}

We presented a controlled comparison between classical and learning-based approaches for real-time motion planning in dynamic hazard fields under a shared formulation. Across four regimes, the results consistently show that deterministic structure favors classical planning, while stochasticity shifts the advantage toward learning-based policies.

In the deterministic regime, classical planners achieve near-perfect success and better path quality, with a clear speed-optimality tradeoff between A* and GBFS. RL offers dramatically lower latency but does not match classical success rates or trajectory quality in this regime. In the stochastic regime, classical performance becomes strongly budget-dependent: increased computation improves success but incurs steep costs in time and produces longer, more conservative paths. RL, in contrast, maintains stable success across visibility settings while preserving very low per-step latency, yielding a favorable success-quality-latency tradeoff under uncertainty.

Overall, our results support a single dominant takeaway: uncertainty in obstacle dynamics, rather than partial observability, is the key factor that changes planning difficulty and determines which paradigm is practically effective in real time.

\bibliographystyle{ieeetr} 
\bibliography{AROB_TexTemplate/refs}

\appendix
\setcounter{section}{0}
\renewcommand{\thesection}{\Alph{section}}
\renewcommand{\thesubsection}{\thesection.\arabic{subsection}}
\renewcommand{\thesubsubsection}{\thesubsection.\arabic{subsubsection}}

\section{Supplementary Material}
\label{sec:extra}

\subsection{Extended RL Formulation and Training Details}
\label{app:rl_details}

\subsubsection{Safety \& Observations.}
Sprinklers are angular sectors; non-flight zones (NFZs) are defined as a uniform inflation of each sector with margin $1$. This encourages safety-aware behavior by penalizing proximity to hazards rather than only collisions.

The agent observes two ego-centered RGB images of size $3\times101\times101$ at scales $0.5$ and $2$. The multi-scale representation enables both local precision and global context.

Agent: $3\times3$ (blue $0.5$); target: $3\times3$ (green $1$); Out-of-Bounds (OOB): blue $1$; sprinklers/NFZ: red channel.  
For constant angular velocity: sprinkler $=1$, NFZ $=0.5$; for variable angular velocity: values scale from $1/3$ to $1$, NFZ is half.

Under limited visibility, only visible sprinklers are rendered; the agent is memoryless.

\subsubsection{Reward \& Curriculum.}
Rewards include goal completion ($25$), NFZ penalty (-$0.25$), progress shaping ($0.005$), and penalties for OOB and collisions (-$2$ each). OOB and collisions terminate the episode.

We employ two curricula: (i) \textbf{termination} - OOB initially does not terminate to avoid degenerate policies, then later it does; (ii) \textbf{complexity} - the number of sprinklers increases from $0$ to $9$-$12$.

\subsubsection{Architecture \& Training.}
Each image is processed by a CNN \citep{mnih2015human}, producing $256$ features; concatenation yields $512$, fed to separate actor and critic MLPs with layers $[128,64]$.

We use $32$ parallel environments and PPO with lr = $5\cdot10^{-5}$, $n_{\text{steps}}=128$, batch = $2,048$, entropy = $0.01$, clip = $0.2$, $\lambda = 0.98$, epochs = $4$, and discount $\gamma = 0.995$.

Training consists of $120$M base steps (full visibility, constant angular velocity), followed by $80$M steps for each regime: full visibility with variable angular velocity, limited visibility with constant angular velocity, and limited visibility with variable angular velocity.

The average wall-clock training throughput was approximately 550 FPS, corresponding to about 2 million environment steps per hour.

\subsection{Experimental Hardware Setup}
\label{app:hardware}
All experiments were conducted on an Azure virtual machine equipped with 8 vCPUs (AMD EPYC 7V12 processor, 2.45 GHz) and 54 GB RAM, under a Microsoft Azure hypervisor. The system also included an NVIDIA Tesla T4 GPU with 16 GB memory. The software environment consisted of Ubuntu 20.04.6 LTS with Python 3.10.13.

The classical planning algorithms (A*, GBFS, and UCT) were executed on the CPU only, each restricted to a single CPU core, without GPU acceleration.

The RL approach was evaluated under two configurations:
(1) CPU-only execution, using a single CPU core, and
(2) GPU-accelerated execution, using a single NVIDIA Tesla T4 GPU.

All experiments were conducted on an otherwise idle virtual machine to minimize variability.

\subsection{A* Offline-Time Histogram}
\label{app:astar_offline_hist}

\begin{figure*}[t]
\centering

\includegraphics[width=0.85\textwidth]{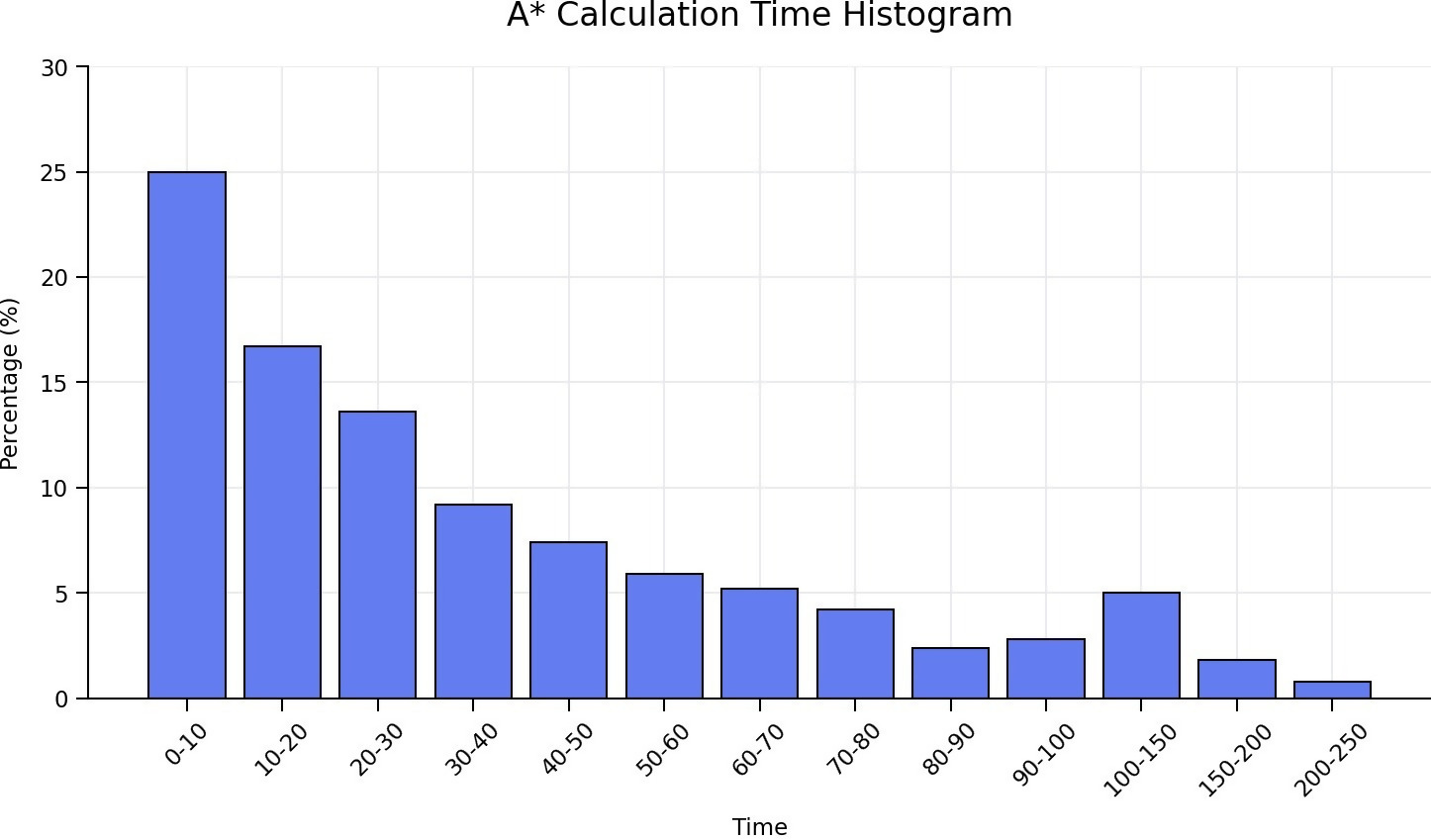}

\caption{
Histogram of A* offline computation time in seconds over the 1,000 deterministic full-visibility scenarios.
The distribution is strongly right-skewed: most runs finish within tens of seconds, while rare hard instances produce a long tail, inflating the standard deviation reported in Table~\ref{tab:constant_speed}.
}
\label{fig:astar_hist}
\end{figure*}

Table~\ref{tab:constant_speed} reports A* offline planning time as mean and standard deviation over 1{,}000 deterministic full-visibility scenarios.
The large standard deviation is explained by a strongly right-skewed, heavy-tailed runtime distribution: most instances are solved quickly, while a small fraction of hard instances require substantially longer search time. For the detailed breakdown, see Fig.~\ref{fig:astar_hist}.

\end{document}